\documentclass{article} 
\usepackage[T1]{fontenc}
\usepackage{iclr2027_conference,times}
\usepackage{graphicx}

\usepackage{amsmath,amsfonts,bm}

\def\eqref#1{equation~\ref{#1}}

\def\1{\bm{1}}

\DeclareMathAlphabet{\mathsfit}{\encodingdefault}{\sfdefault}{m}{sl}
\SetMathAlphabet{\mathsfit}{bold}{\encodingdefault}{\sfdefault}{bx}{n}

\usepackage{multirow}
\usepackage{hyperref}
\usepackage{url}
\usepackage{booktabs}
\usepackage{wrapfig}

\newcommand{\methodname}{MERA}

\title{Learning to Retrieve Missing Evidence for Long-Term Memory QA}

\author{Yi-Xuan Deng\textsuperscript{1}, Yi Zhang\textsuperscript{1}, Wei Liu\textsuperscript{2}, Chao Xue\textsuperscript{2}, Shuojin Yang\textsuperscript{1}\\
\textsuperscript{1}Tsinghua University \quad \textsuperscript{2}JD.COM}

\iclrfinalcopy
\begin{document}

\maketitle
\lhead{Preprint}
\begin{abstract}
Long-term memory enables language models to use past interactions in future conversations. However, evidence needed to answer a question may be scattered across distant turns, while the question itself omits clues needed to locate it. Retrieved facts can reveal these clues, motivating retrieval decisions conditioned on evidence already found. We introduce \methodname{} (Missing-Evidence Retrieval Augmentation), which separates globally searchable memory from a question-specific evidence state. Verified evidence guides subsequent retrieval without restricting access to the global memory. We train a lightweight planner through reinforcement learning, rewarding queries that recover previously missing evidence. \methodname{} achieves strong answer accuracy across Qwen3-30B and GPT-4o-mini backbones. With Qwen3-30B for evidence processing and answer generation, the trained 0.6B planner achieves 77.40\% accuracy on LoCoMo and 71.29\% on LongMemEval-S, exceeding a 30B planner without retrieval-grounded training by 4.10\% and 3.96\%, respectively. On LoCoMo, later retrieval rounds increase cumulative evidence recall from 55.5\% to 80.5\%.
\end{abstract}


\section{Introduction}
\label{sec:introduction}

Long-term memory enables language models to retain information from past interactions and use it in future conversations~\cite{MemoryBank,GenerativeAgents}. Recent memory systems have made substantial progress in deciding what to store, how to organize it, and how to maintain it over time~\cite{MemorySurvey,LightMem}. However, answering a question over long conversational histories poses a different challenge: evidence is often sparse and distributed across distant turns, and the question itself may omit the entities, events, or temporal cues needed to locate it~\cite{Locomo,LongMem-eval}. As a result, retrieving memories individually relevant to the question may still leave critical evidence undiscovered.

The key challenge is therefore not only which memories are relevant, but \emph{what should be retrieved next given what has already been found}~\cite{Survey-RetrievalAugmented-LLM}. Recovered evidence can supply search anchors absent from the original question~\cite{EfficientRAG}. Persistent-memory methods primarily maintain reusable memory, with less emphasis on a question-specific state of verified evidence~\cite{Mem0,A-mem,MemoryOS}. IRCoT~\cite{IRCoT} adapts queries using accumulated passages and reasoning, without an explicit verification step separating established evidence from other intermediate content. In methods that retrieve through graph traversal, access to further evidence depends partly on the preconstructed graph's connectivity~\cite{HippoRAG,KnowledgeGraphs,GraphReader,Think-on-Graph}.

\begin{figure}[t]
    \centering
    \includegraphics[width=\linewidth]{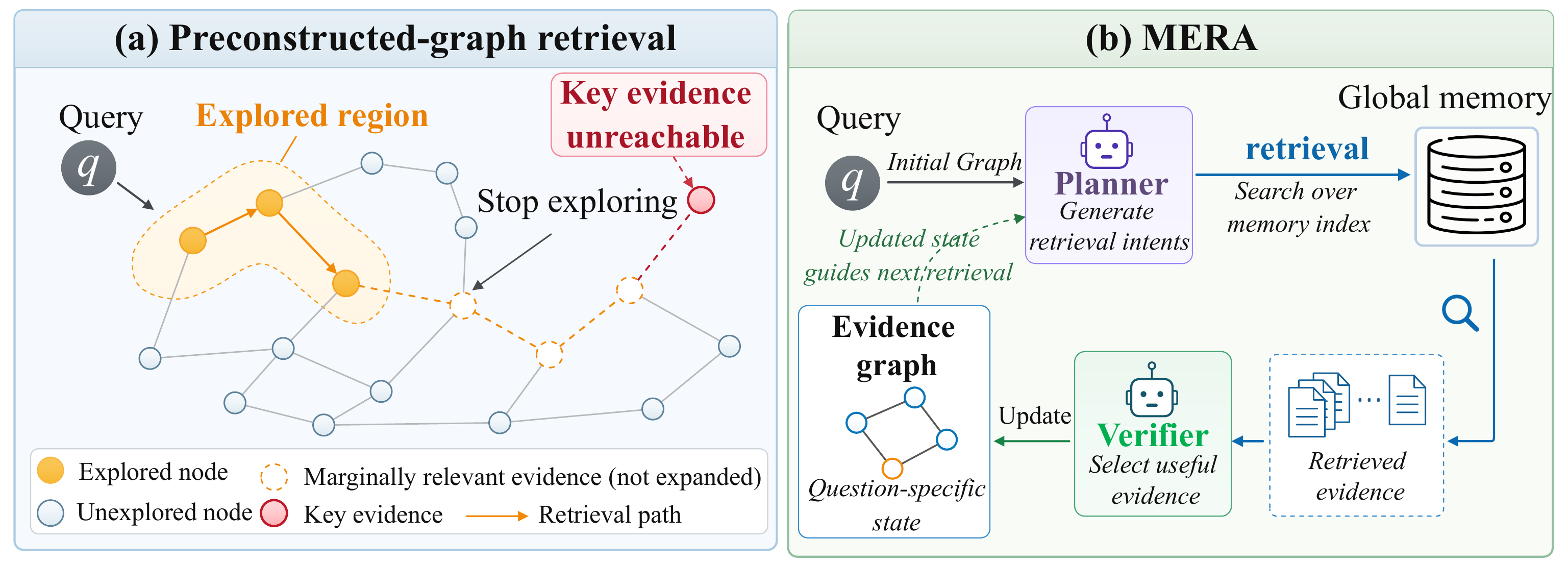}
    \caption{A question-specific evidence state should guide retrieval without defining the search boundary.
    (a) A preconstructed graph's explored region may exclude relevant evidence.
    (b) \methodname{} searches a separate global memory index; verified evidence updates the state and guides subsequent retrieval.
    Orange and green diamonds mark required and verified evidence; panels are schematic.}
    \label{fig:motivation}
\end{figure}

We introduce \methodname{} (Missing-Evidence Retrieval Augmentation), which separates the question-specific evidence state from the global search space. The evidence graph organizes established facts within the planner's information state, helping identify what remains missing. As Figure~\ref{fig:motivation} illustrates, newly established entities or temporal relations can guide retrieval over the global index, including evidence not yet represented in the graph. \methodname{} pairs a global atomic-fact index with source-turn identifiers~\cite{DenseX-Retrieval} and an evidence graph for each question. A lightweight planner generates structured retrieval intents from the graph, question, and query history. A verifier gates which retrieved facts enter the graph to inform subsequent retrieval~\cite{Survey-LLM-Agent,ReAct}. Retrieval continues until the evidence is judged sufficient, the search is exhausted, or the round budget is reached. The final graph and accepted evidence then provide the context for answer generation.

We train the planner to recover new evidence rather than reproduce a prescribed query. Differently worded queries may retrieve the same evidence, while a plausible query may only repeat earlier hits~\cite{Survey-retrieval}. We therefore evaluate actions by their retrieval outcomes~\cite{CONQRR,Query-Rewriting}: for each intermediate evidence state, we execute multiple planner outputs against the memory index and reward newly recovered gold evidence, with no credit for repeated hits. Comparing queries from the same evidence state provides relative feedback on how effectively each action closes the remaining evidence gap. These rewards train the lightweight planner with Group Relative Policy Optimization (GRPO)~\cite{DeepSeekMath}, while the memory index, evidence model, and answer model remain fixed.

\methodname{} achieves strong answer accuracy across Qwen3-30B and GPT-4o-mini backbones. Using Qwen3-30B~\cite{Qwen3} for evidence processing and answer generation, \methodname{}'s retrieval-trained 0.6B planner achieves 77.40\% accuracy on LoCoMo and 71.29\% on LongMemEval-S. It also exceeds a 30B planner without retrieval-grounded training by 4.10 and 3.96 percentage points, respectively. On LoCoMo, later retrieval rounds raise cumulative evidence recall from 55.5\% to 80.5\%.

Our contributions are threefold:
\begin{itemize}
    \item We identify a query-time gap in long-term memory QA: persistent memory describes what can be retrieved, but answering a question additionally requires tracking what has been established and deciding what to retrieve next.

    \item We introduce \methodname{}, which decouples a verifier-gated evidence graph from a globally searchable atomic-fact index, allowing verified evidence to guide search without restricting access to other relevant evidence.

    \item We train a lightweight planner using retrieval outcomes, rewarding previously missing evidence rather than supervising a unique query form. Under the evaluated settings, the trained 0.6B planner achieves higher answer accuracy than a 30B planner without retrieval-grounded training.
\end{itemize}

\section{Related Work}
\label{sec:related-work}

\subsection{Long-Term Memory for LLM Agents}

Long-term memory systems mainly optimize persistent representation~\cite{MemorySurvey}. Mem0~\cite{Mem0} maintains salient facts through explicit add, update, delete, and no-op decisions; A-MEM~\cite{A-mem} organizes experiences as interconnected notes that evolve with new memories. MemoryOS~\cite{MemoryOS} manages recent dialogue, topic-level pages, and personal knowledge in separate tiers, whereas LightMem~\cite{LightMem} compresses and consolidates interactions offline. These methods operate primarily on a store shared across future questions. \methodname{} focuses on the retrieval decisions made while answering a particular question: it keeps the memory index fixed and maintains an explicit evidence state that lets the planner use verified facts and previous searches to identify missing information and decide what to retrieve next.

\subsection{Learning to Manage and Use Memory}

Memory-R1~\cite{Memory-R1} applies reinforcement learning to persistent-memory operations and to selecting stored entries for answering. Lightweight LLM Agent Memory~\cite{lightweightLLM} uses small models to coordinate retrieval, writing, and consolidation under a fixed retrieval budget, while MemSkill~\cite{MemSkill} learns to select and evolve reusable memory-construction skills. \methodname{} optimizes the query planner at intermediate evidence states, using the previously missing gold evidence turns recovered by each executed action as its primary reward. The learned action specifies what to search for next, conditioned on what earlier retrieval rounds have established; the persistent memory, evidence model, and answer model remain fixed during training.

\subsection{Iterative Retrieval and Structured Evidence}

IRCoT~\cite{IRCoT} alternates retrieval with chain-of-thought generation, using the latest reasoning sentence as the next query and accumulated passages and thoughts as its search context. EfficientRAG~\cite{EfficientRAG} trains lightweight token classifiers to retain useful passage tokens and assemble next-hop queries. FLARE~\cite{FLARE} retrieves from low-confidence anticipated content, while Adaptive-RAG~\cite{adaptiveRAG} routes questions among no-retrieval, single-step, and iterative strategies. These methods establish the value of adapting retrieval during answering~\cite{Survey-RetrievalAugmented-LLM}. \methodname{} addresses the next-query decision through an explicit, verifier-gated evidence state and trains the planner according to the missing evidence recovered by its actions, favoring queries that complement the facts already verified in earlier rounds.

GraphReader~\cite{GraphReader} constructs a document graph of key elements and atomic facts, then explores selected nodes, neighbors, and source chunks. Think-on-Graph~\cite{Think-on-Graph} searches an existing knowledge graph by iteratively selecting relations and entities with beam search, while HippoRAG~\cite{HippoRAG} propagates query signals over a persistent corpus graph. These methods search graphs constructed independently of the current question. \methodname{} instead retrieves from a fixed atomic-fact index and incrementally builds a transient graph from verified, source-linked evidence. The graph organizes the planner state without making evidence access depend on preconstructed graph connectivity, allowing subsequent queries to retrieve evidence about entities not yet represented in the graph or linked to its existing nodes.

\section{Method}
\label{sec:method}

\begin{figure}[t]
    \centering
    \includegraphics[width=0.96\linewidth]{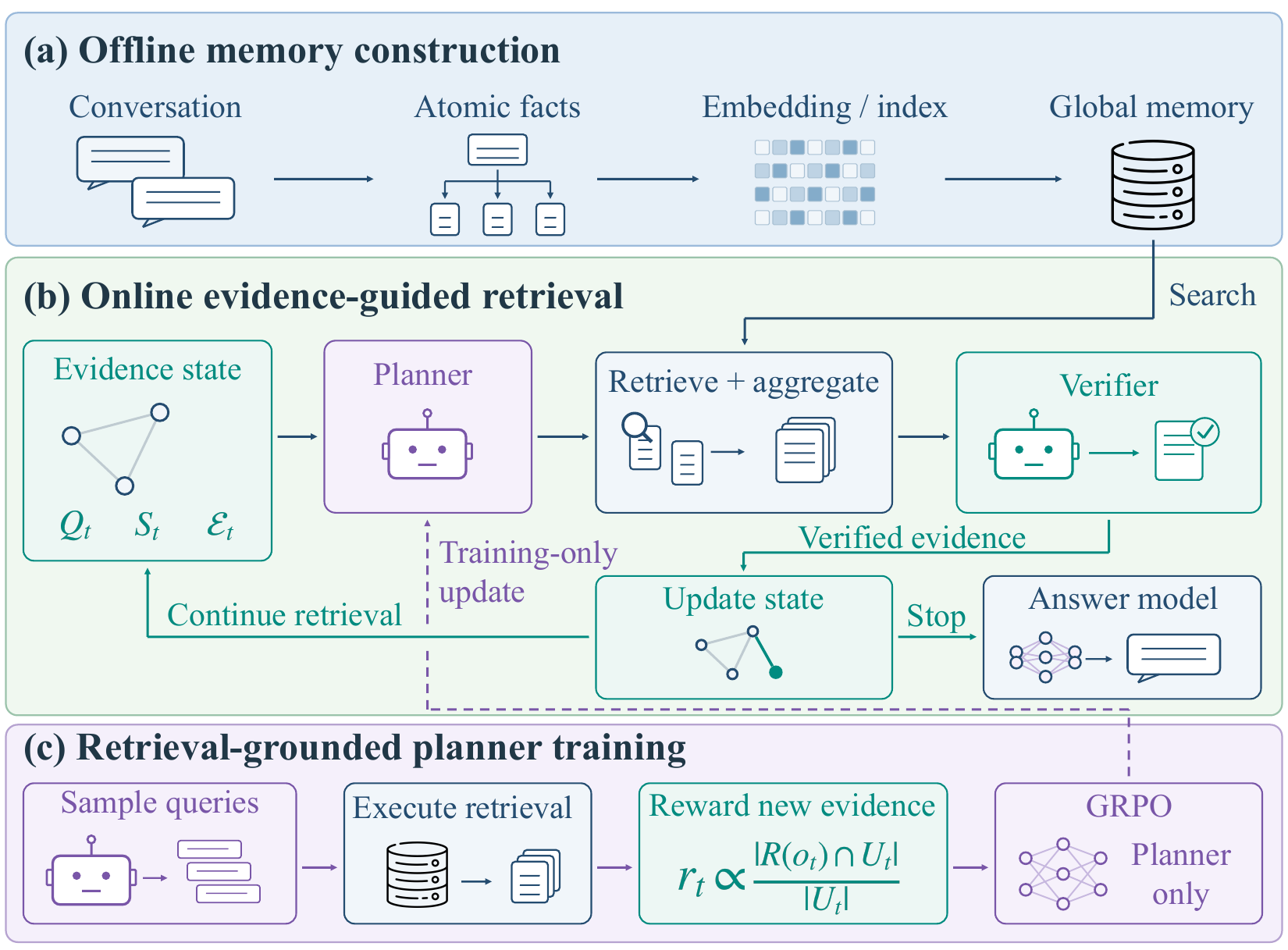}
    \caption{Overview of \methodname{}.
    (a) Conversation turns are decomposed into atomic facts and indexed in global memory.
    (b) Planning, retrieval, aggregation, and verification update the evidence state for continued retrieval or answering.
    (c) Newly recovered gold evidence provides the reward for GRPO; the dashed arrow denotes the training-only planner update.}
    \label{fig:pipeline}
\end{figure}

\subsection{Problem Formulation}
\label{sec:problem-formulation}

Let a long-term conversation history be a chronological sequence of turns $\mathcal{U}=\{u_1,\ldots,u_N\}$, where each turn includes its utterance, speaker, timestamp, and session metadata. Given a question $q$, the goal is to produce an answer supported by a sparse subset of turns. Because an entity or relation discovered in one turn may reveal what to retrieve next, the original question is often insufficient to recover the complete evidence chain in a single search.

We therefore formulate long-term memory QA as sequential evidence acquisition, where each retrieval decision depends on what has already been established and what the question still requires. At round $t$, a planner generates retrieval actions $a_t$ from a question-specific evidence state $s_t$; a retriever executes them against persistent memory $\mathcal{M}$; and a verifier accepts useful evidence and updates the state:

\begin{equation}
a_t\sim\pi_\theta(\cdot\mid s_t),\quad
C_t=\operatorname{Retrieve}(\mathcal{M},a_t),\quad
(E_t,\Delta G_t,b_t)=V_\phi(q,G_t,\mathcal{E}_t,C_t).
\end{equation}

Here, $C_t$ contains retrieved candidates, $E_t$ is newly accepted evidence, $\Delta G_t$ is the graph update, and $b_t$ indicates whether the evidence is sufficient. The objective is to acquire enough evidence for a correct answer within a bounded retrieval budget.

\subsection{\methodname{} Overview}
\label{sec:gara-overview}

\methodname{} separates reusable memory from the evidence state maintained for each question through three stages (Figure~\ref{fig:pipeline}). Offline, conversation turns are decomposed into self-contained atomic facts, embedded, and indexed with parent-turn identifiers. Online, a lightweight planner guides global retrieval using the evolving evidence state; parent-turn fact aggregation and verification supply evidence for further search or answering. As verified facts accumulate, they provide new entities and relations for subsequent queries, while global retrieval keeps evidence outside the current graph accessible. During training, multiple queries sampled from a shared state are executed against the same memory and rewarded for recovering previously missing gold evidence, giving the planner feedback on the actual retrieval outcomes. GRPO updates only the planner, while the memory index, evidence model, and answer model remain fixed.

\subsection{Traceable Memory and Evidence State}
\label{sec:atomic-memory}
\label{sec:evidence-graph}

For each conversation turn, \methodname{} extracts self-contained atomic facts with retrieval metadata and embeds them for matching. A matched fact provides a focused similarity signal but may omit related information captured in other facts from the same turn, so before verification \methodname{} aggregates all extracted facts belonging to each matched parent turn, together with its speaker, timestamp, and source identifier. We call this operation \emph{parent-turn fact aggregation}. Results from multiple intents are merged and deduplicated by parent turn, giving the verifier a broader fact record, containing extracted facts rather than verbatim source text, while retaining provenance.

For question $q$, \methodname{} represents the accumulated evidence in a transient graph $G_t$, within the complete evidence state $s_t$:

\begin{equation}
G_t=(V_t,L_t,X_t,D_t,\Psi_t),\qquad
s_t=(q,G_t,Q_t,S_t,\mathcal{E}_t).
\end{equation}

The graph stores verified entities and relations $(V_t,L_t)$, active retrieval anchors $X_t$, exhausted or invalid directions $D_t$, and provenance links $\Psi_t$ to source turns. Alongside the graph, the evidence-state box in Figure~\ref{fig:pipeline}(b) shows previous queries $Q_t$, viewed candidates $S_t$, and accepted evidence $\mathcal{E}_t$. Together, these records let the planner compare what has been established and attempted against the question to identify what remains missing. The graph is constructed for the current question, while the persistent index determines which memories remain searchable. Appendices~\ref{app:atomic-memory} and~\ref{app:evidence-graph} detail atomic-fact construction, parent-turn fact aggregation, graph merging, and temporal normalization.

\subsection{Evidence-State-Guided Active Retrieval}
\label{sec:active-retrieval}

The evidence state is initialized from the question. At round $t$, the planner compares the information required by $q$ with the current evidence state. If a gap remains, it emits a small set of structured retrieval intents,

\begin{equation}
a_t=\{I_t^\ell\}_{\ell=1}^{m_t},\qquad
I_t^\ell=(d_t^\ell,k_t^\ell,c_t^\ell),
\end{equation}

where $d_t^\ell$ describes the desired fact, $k_t^\ell$ identifies salient entities or concepts, and $c_t^\ell$ supplies coarse topical labels. Conditioning on verified evidence and query history lets later intents use newly discovered anchors and avoid repeating earlier searches.

The retriever matches each intent against the global atomic-fact index, assembles the corresponding parent-level fact records, and removes previously viewed evidence. Candidate access does not require adjacency to a node in $G_t$: a semantic query can retrieve a fact about an entity absent from the current graph. The verifier admits only candidates that help resolve the question and updates the state as

\begin{equation}
G_{t+1}=\operatorname{Merge}(G_t,\Delta G_t),\quad
\mathcal{E}_{t+1}=\mathcal{E}_t\cup E_t,\quad
Q_{t+1}=Q_t\cup a_t.
\end{equation}

The green node and edge in the \emph{Update state} box illustrate newly verified information incorporated into the graph. The \emph{Continue retrieval} branch returns the updated state to the planner. When evidence is sufficient, search is exhausted, or the round budget is reached, the \emph{Stop} branch passes the final graph and accepted fact records, with their source identifiers, to the answer model. Appendix~\ref{app:active-retrieval} specifies the retrieval and termination rules; Appendix~\ref{app:rl-trajectory} provides a logged trajectory, and Appendix~\ref{app:runtime-statistics} reports planner runtimes and the inference environment.

\subsection{Retrieval-Grounded Policy Optimization}
\label{sec:policy-optimization}

Query generation does not have a unique correct string, so \methodname{} evaluates a planner output by executing it against the actual memory. Planner-training questions come from dataset-specific sources: LoCoMo questions are synthesized with the benchmark's released data-generation utilities, while LongMemEval questions are generated from the designated training conversations. Training states are collected before planner calls throughout complete retrieval trajectories, spanning initial search, complementary retrieval, and late-stage completion; states with uniformly unsuccessful or uniformly complete outcomes are removed so optimization focuses on decisions where the query can change retrieval results. Appendix~\ref{app:rl-data-construction} details the data construction process.

Let $\mathcal{E}^*$ be the gold parent turns, $F_t$ the gold evidence found before round $t$, $U_t=\mathcal{E}^*\setminus F_t$ the remaining evidence, and $R(o_t)$ the parent turns retrieved by planner output $o_t$. We define

\begin{equation}
x_t=\frac{|R(o_t)\cap U_t|}{|U_t|},\qquad
p_t=\frac{|F_t|}{|\mathcal{E}^*|},\qquad
R_{\mathrm{ret}}(o_t)=x_t\left[1+p_t(1-x_t)\right].
\end{equation}

The fraction shown in Figure~\ref{fig:pipeline}(c) is $x_t$, the proportion of missing gold evidence recovered by an action. The figure's $r_t\propto x_t$ is schematic: the implemented retrieval reward $R_{\mathrm{ret}}$ also includes the progress term above. Evidence already in $F_t$ receives no additional credit, and queries with different wording receive the same retrieval reward when they recover the same missing gold turns. The progress term increases the reward for partial recovery of remaining evidence later in a trajectory, and parent-turn-level reward aligns training with gold annotations, verification, and answer generation. Auxiliary keyword-count and response-length penalties shape the planner output, as described in Appendix~\ref{app:policy-optimization}.

We optimize the lightweight planner with Group Relative Policy Optimization (GRPO)~\cite{DeepSeekMath}: for each shared evidence state, \methodname{} executes a group of sampled outputs, normalizes the resulting rewards into relative advantages, and applies the clipped policy objective with KL regularization and an optional entropy term~\cite{PPO}, so that with a shared question, graph, and memory across the group, relative rewards primarily reflect the sampled query actions. Appendix~\ref{app:policy-optimization} gives the complete trajectory fields, filtering rule, reward rationale, and a detailed derivation of the GRPO objective.

\section{Experiments}
\label{sec:experiments}

We evaluate whether deciding what to retrieve next from accumulated evidence improves long-term memory QA. After reporting answer accuracy and token costs, we examine whether later queries recover missing evidence, how the evidence state supports retrieval and answering, and whether retrieval-grounded training improves the lightweight planner. The appendix examines parent-turn fact aggregation, query behavior, runtime, termination, and retrieval-hyperparameter sensitivity.

\subsection{Experimental Setup}
\label{sec:experimental-setup}

We use LoCoMo~\cite{Locomo}, which contains 1,986 questions, and LongMemEval-S~\cite{LongMem-eval}, which contains 500 questions. The compared systems are FullText, NaiveRAG~\cite{RAG}, IRCoT~\cite{IRCoT}, GraphReader~\cite{GraphReader}, LangMem~\cite{langchain2025langmem}, A-MEM~\cite{A-mem}, MemoryOS~\cite{MemoryOS}, Mem0~\cite{Mem0}, LightMem~\cite{LightMem}, and \methodname{}-RL. We report results with Qwen3-30B-A3B-Instruct-2507~\cite{Qwen3} and GPT-4o-mini~\cite{gpt4omini} as the cache-construction and inference backbone.

Answer accuracy is measured by an LLM judge. Generative-model costs include offline memory construction and online question answering but exclude embedding and vector retrieval. Appendix~\ref{app:experimental-setup} provides dataset descriptions and splits, metric accounting, and implementation details.

\subsection{Main Results}
\label{sec:main-results}

Table~\ref{tab:main-results} jointly compares answer accuracy and offline memory-construction cost on LongMemEval-S and LoCoMo. We use three \methodname{} configurations throughout the experiments. \textbf{\methodname{}-Small} uses a Qwen3-0.6B query planner without retrieval-grounded training, while \textbf{\methodname{}-RL} uses the same planner after retrieval-grounded GRPO~\cite{DeepSeekMath} training. In both configurations, the evidence model is the backbone specified in the corresponding table block. \textbf{\methodname{}-Large} uses that table-specified backbone for both query planning and evidence processing, without retrieval-grounded planner training.

\stepcounter{footnote}
\begin{table}[t]
\caption{Main results on LongMemEval-S and LoCoMo. ACC is answer accuracy. Offline generative-model costs are reported in thousands per question (k/Q) for LongMemEval-S and thousands per conversation (k/Conv.) for LoCoMo. Superscript $^{\ast}$ marks results obtained from our own runs.\protect\footnotemark[\value{footnote}] ``--'' denotes not applicable or not reported; IRCoT makes no offline generative calls.}
\label{tab:main-results}
\label{tab:lmem-main}
\label{tab:locomo-main}
\centering
\scriptsize
\setlength{\tabcolsep}{3pt}
\begin{tabular*}{\textwidth}{@{\extracolsep{\fill}}llrrrrrr@{}}
\hline
\rule{0pt}{2.4ex}\textbf{Backbone} & \textbf{Method} & \textbf{ACC (\%)} & \textbf{Sum. In (k)} & \textbf{Sum. Out (k)} & \textbf{Upd. In (k)} & \textbf{Upd. Out (k)} & \textbf{Total (k)} \\[0.8ex]
\hline
\multicolumn{8}{c}{\rule{0pt}{2.2ex}\textbf{LongMemEval-S}} \\[0.6ex]
\hline
\multirow{10}{*}{\textbf{Qwen3-30B}} & FullText & 54.80 & -- & -- & -- & -- & 105.07 \\
& NaiveRAG & 60.80 & -- & -- & -- & -- & -- \\
& IRCoT$^{\ast}$ & 55.45 & -- & -- & -- & -- & -- \\
& GraphReader$^{\ast}$ & 46.53 & 155.52 & 99.61 & -- & -- & 255.13 \\
& LangMem & 50.80 & -- & -- & 1,311.96 & 118.06 & 1,430.02 \\
& A-MEM & 65.20 & 219.21 & 66.98 & 1,260.54 & 318.20 & 1,864.93 \\
& MemoryOS & 49.60 & 2,101.54 & 510.88 & 305.12 & 27.43 & 2,944.97 \\
& Mem0 & 39.51 & 424.20 & 15.34 & 411.50 & 111.35 & 962.39 \\
& LightMem$^{\ast}$ & 67.33 & 13.19 & 19.21 & -- & -- & 32.40 \\
& \textbf{\methodname{}-RL}$^{\ast}$ & \textbf{71.29} & 328.74 & 91.97 & -- & -- & 420.70 \\
\hline
\multirow{8}{*}{\textbf{GPT-4o-mini}} & FullText & 56.80 & -- & -- & -- & -- & 105.07 \\
& NaiveRAG & 61.00 & -- & -- & -- & -- & -- \\
& LangMem & 37.20 & -- & -- & 982.68 & 119.48 & 1,102.16 \\
& A-MEM & 62.60 & 214.66 & 42.82 & 1,157.52 & 190.81 & 1,605.81 \\
& MemoryOS & 44.80 & 2,302.35 & 304.18 & 350.02 & 35.19 & 2,991.74 \\
& Mem0 & 53.61 & 424.13 & 17.76 & 560.17 & 150.56 & 1,152.62 \\
& LightMem$^{\ast}$ & 63.37 & 31.65 & 11.25 & -- & -- & \textbf{42.91} \\
& \textbf{\methodname{}-RL}$^{\ast}$ & \textbf{65.35} & 313.83 & 83.23 & -- & -- & 397.07 \\
\hline\hline
\multicolumn{8}{c}{\rule{0pt}{2.2ex}\textbf{LoCoMo}} \\[0.6ex]
\hline
\multirow{11}{*}{\textbf{Qwen3-30B}} & FullText & 74.87 & -- & -- & -- & -- & -- \\
& NaiveRAG & 66.95 & -- & -- & -- & -- & -- \\
& IRCoT$^{\ast}$ & 62.34 & -- & -- & -- & -- & -- \\
& GraphReader$^{\ast}$ & 75.18 & 59.24 & 37.99 & -- & -- & 97.24 \\
& LangMem & 60.53 & -- & -- & 1,004.35 & 138.02 & 1,142.37 \\
& A-MEM & 56.10 & 158.29 & 60.85 & 924.19 & 483.51 & 1,626.84 \\
& MemoryOS (LoCoMo) & 61.04 & 122.21 & 53.12 & 104.43 & 81.75 & 361.51 \\
& MemoryOS (regular) & 51.30 & 228.85 & 51.60 & 242.27 & 143.63 & 666.35 \\
& Mem0 & 43.31 & 827.09 & 18.64 & 763.88 & 189.80 & 1,799.41 \\
& LightMem$^{\ast}$ & 61.58 & 61.38 & 36.33 & 9.86 & 0.88 & 108.45 \\
& \textbf{\methodname{}-RL}$^{\ast}$ & \textbf{77.40} & 148.32 & 19.73 & -- & -- & 168.05 \\
\hline
\multirow{9}{*}{\textbf{GPT-4o-mini}} & FullText & \textbf{71.83} & -- & -- & -- & -- & -- \\
& NaiveRAG & 63.64 & -- & -- & -- & -- & -- \\
& LangMem & 57.20 & -- & -- & 898.27 & 111.95 & 1,010.22 \\
& A-MEM & 64.16 & 182.74 & 49.29 & 729.89 & 187.52 & 1,149.44 \\
& MemoryOS (LoCoMo) & 58.25 & 110.98 & 33.40 & 78.08 & 64.54 & 287.00 \\
& MemoryOS (regular) & 54.87 & 226.86 & 46.61 & 177.66 & 75.34 & 526.47 \\
& Mem0 & 61.69 & 851.32 & 20.53 & 632.12 & 189.42 & 1,693.39 \\
& LightMem$^{\ast}$ & 63.29 & \textbf{62.82} & \textbf{17.95} & 4.14 & 0.28 & \textbf{85.19} \\
& \textbf{\methodname{}-RL}$^{\ast}$ & \textbf{67.02} & 270.07 & 36.79 & -- & -- & 306.86 \\
\hline
\end{tabular*}
\end{table}

\methodname{}-RL achieves the highest answer accuracy among the compared methods on both benchmarks with Qwen3-30B. With GPT-4o-mini, it also achieves the highest accuracy on LongMemEval-S and the strongest result among the memory-based systems on LoCoMo. The results show that \methodname{} can recover and use relevant conversational evidence across both datasets and backbone settings.

This performance is consistent with the way \methodname{} connects evidence acquisition to the current question state. Atomic facts provide focused retrieval targets, and parent-turn fact aggregation supplies related details for verification. The evidence graph then connects accepted facts across sessions and exposes entities, relations, and temporal anchors that can guide subsequent queries. As evidence accumulates, the planner can refine its search around information still needed to answer the question. Global retrieval keeps evidence accessible even when it concerns an entity absent from the current graph. These properties are useful for long histories, where the initial question may omit the intermediate clues needed to locate supporting evidence. Section~\ref{sec:evidence-coverage} and the ablations below examine this process through evidence recovery and component comparisons.
\footnotetext[\value{footnote}]{Results marked with $^{\ast}$ are obtained from our own runs under the evaluation setup described above, including our IRCoT, GraphReader, LightMem, and \methodname{}-RL evaluations.}

\methodname{} combines this accuracy with lower offline token consumption than most memory systems in Table~\ref{tab:main-results}. Its construction cost is concentrated in extracting atomic facts from conversation turns; the resulting index is reusable across questions. The pipeline processes each turn without repeatedly presenting an expanding memory to a generative model for consolidation or rewriting. Question-specific evidence organization takes place during retrieval, so the offline stage need not anticipate every relation that a future question may require. This division of work helps limit memory-maintenance overhead while retaining fine-grained evidence for later search.

\subsection{Does Retrieval Feedback Improve a Lightweight Planner?}
\label{sec:planner-analysis}

We compare \methodname{}-Small (0.6B) and \methodname{}-Large (30B), both without retrieval-feedback training, with \methodname{}-RL, the trained 0.6B planner. All three use Qwen3-30B for evidence processing and answering; we measure answer accuracy.

\subsubsection{Planner Scale and Policy-Training Ablation}

\begingroup
\setlength{\intextsep}{3pt}
\begin{wraptable}{l}{0.49\textwidth}
\setlength{\abovecaptionskip}{0pt}
\setlength{\belowcaptionskip}{4pt}
\caption{Planner-scale and policy-training ablation: answer accuracy (\%).}
\label{tab:scale-rl-ablation}
\centering
\normalsize
\setlength{\tabcolsep}{3pt}
\renewcommand{\arraystretch}{1.15}
\begin{tabular*}{\linewidth}{@{\extracolsep{\fill}}lcc@{}}
\hline
\textbf{Planner} & \textbf{LoCoMo} & \textbf{LongMemEval-S} \\
\hline
\methodname{}-Small & 68.40 & 59.80 \\
\methodname{}-Large & 73.30 & 67.33 \\
\methodname{}-RL & \textbf{77.40} & \textbf{71.29} \\
\hline
\end{tabular*}
\end{wraptable}

Table~\ref{tab:scale-rl-ablation} shows that both training and scaling improve answer accuracy. Relative to \methodname{}-Small, \methodname{}-Large gains 4.90 points on LoCoMo and 7.53 points on LongMemEval-S, while \methodname{}-RL gains 9.00 and 11.49 points, respectively. The trained 0.6B planner therefore exceeds \methodname{}-Large by 4.10 and 3.96 points. The training objective rewards recovery of evidence missing from the current state, encouraging queries that complement earlier findings. These results show that, under the shared evidence and answer models, training the lightweight planner yields larger accuracy gains than increasing planner size alone.
\par\endgroup

\paragraph{Additional analyses.}
Appendix~\ref{app:planner-diagnostics} examines query behavior and parent-turn fact aggregation. Appendix~\ref{app:error-analysis} analyzes question types and failures, while Appendix~\ref{app:retrieval-sensitivity} reports retrieval-hyperparameter sensitivity.

\subsection{Online Inference Cost}
\label{sec:online-cost}

With Qwen3-30B, \methodname{}'s online token consumption is on the order of tens of thousands of tokens per question and remains below the graph-navigation baseline on both datasets. This cost accompanies the leading answer accuracy reported in Table~\ref{tab:main-results}. The online computation is directed toward successive planning and verification over retrieved evidence. Atomic retrieval selects candidate records, verification filters their relevance, and the evolving graph organizes accepted facts for subsequent queries and final answering. This process focuses model context on evidence related to the question, while allowing additional retrieval when the current evidence is incomplete. Appendix~\ref{app:online-cost} provides the input/output token breakdown for each dataset.

\subsection{How Does Multi-Round Retrieval Improve Evidence Coverage?}
\label{sec:evidence-coverage}

\methodname{} is designed to use the current evidence state to guide follow-up retrieval for information missed in earlier rounds. We measure how much additional gold evidence is recovered after the first round by tracking cumulative turn-level recall throughout retrieval.

\begin{figure}[t]
\centering
\includegraphics[width=\linewidth]{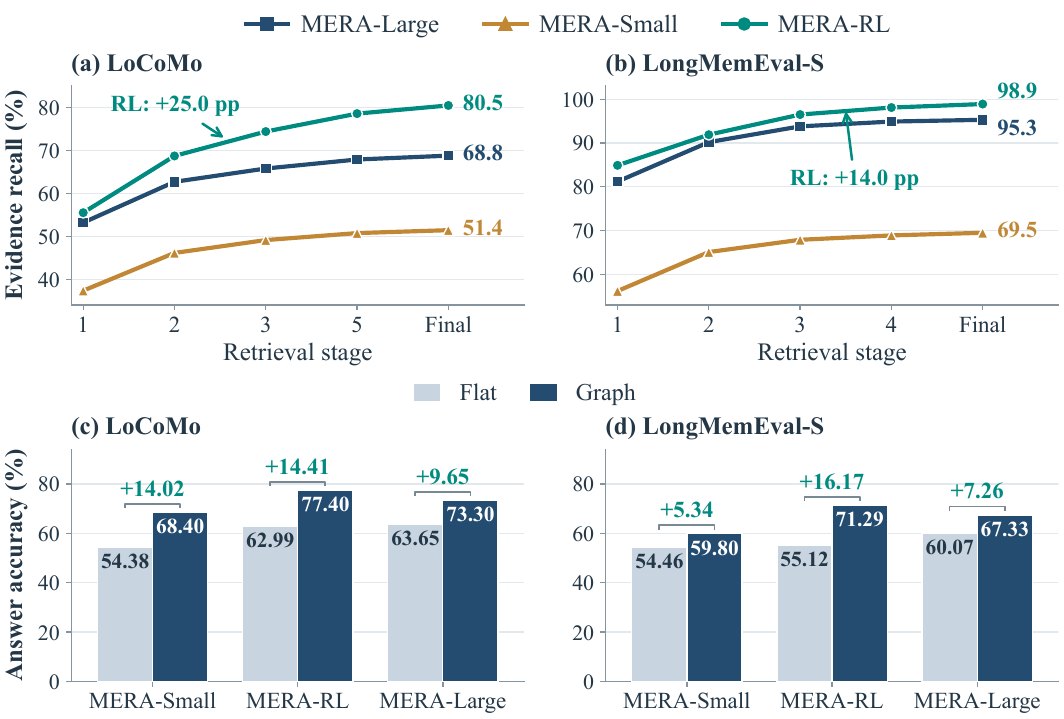}
\caption{Evidence recall and graph ablations. (a)--(b) Cumulative recall with dataset-specific y-axis ranges; arrows show RL gains from round 1 to termination. (c)--(d) Answer accuracy; brackets show Graph minus Flat. All gains are in percentage points.}
\label{fig:evidence-recall}
\label{fig:graph-ablation}
\end{figure}

\paragraph{Later rounds recover complementary evidence.}
On LoCoMo, recall after the first round increases by 14.1, 25.0, and 15.6 percentage points for \methodname{}-Small, \methodname{}-RL, and \methodname{}-Large, respectively. \methodname{}-RL rises from 0.555 to 0.805, showing that later actions recover substantial complementary evidence. On LongMemEval-S, \methodname{}-RL rises from 0.849 to 0.989, a further 14.0-point gain. Among questions with a hit, the first gold evidence is recovered at round 1.13 on average, so additional rounds primarily complete the evidence set. These results support complementary retrieval; Section~\ref{sec:graph-ablation} examines the contribution of the graph-based reasoning process.

Most evidence gain occurs early: at least 92\% of the LoCoMo curves' post-first-round improvement is realized by round 5. Appendix~\ref{app:retrieval-depth-interface} discusses diminishing returns and retriever compatibility; Appendix~\ref{app:termination-behavior} analyzes stopping behavior.

\subsection{How Does the Evidence State Support Retrieval and Answering?}
\label{sec:graph-ablation}

We compare the graph-assisted evidence workflow with a flat-evidence variant that removes the evidence graph and its associated state-maintenance mechanisms. Accepted evidence accumulates in a flat list for planning and final answering; atomic retrieval and parent-turn fact aggregation remain enabled. All no-graph configurations use Qwen3-30B for evidence processing and answering. The no-graph runs use dense atomic-fact retrieval with MMR ($\lambda=0.95$), $k=8$, and a budget of ten rounds. The planners are the base 0.6B model, its retrieval-trained counterpart, and the 30B model. Figure~\ref{fig:graph-ablation}(c)--(d) compares these workflows, rather than isolating graph representation alone. Appendix~\ref{app:graph-ablation} reports the experimental configuration and results.

Graph-assisted configurations attain higher accuracy in all six comparisons. The graph--flat gaps for \methodname{}-Small, \methodname{}-RL, and \methodname{}-Large are 14.02, 14.41, and 9.65 points on LoCoMo, respectively; the corresponding gaps on LongMemEval-S are 5.34, 16.17, and 7.26 points. All graph scores use the corresponding planner configurations and answer accuracies reported in Table~\ref{tab:scale-rl-ablation}. On LoCoMo, the Small and RL gaps are similar, whereas on LongMemEval-S the RL gap is larger than those of the other two planners. The magnitude of the difference therefore varies across both datasets and planner configurations. Positive gaps occur for both planners without retrieval-feedback training as well as the RL planner, so the observed pattern is not confined to the trained policy.

The planner ranking also depends on the evidence workflow. With flat evidence, \methodname{}-Large exceeds \methodname{}-RL by 0.66 points on LoCoMo and 4.95 points on LongMemEval-S. With the graph-assisted workflow, \methodname{}-RL instead leads by 4.10 and 3.96 points. Thus, the trained lightweight planner's advantage over the large planner in Table~\ref{tab:scale-rl-ablation} is observed with the graph-assisted workflow and does not persist in this flat-evidence variant. This comparison characterizes the performance of the evaluated configurations without assigning the differences to a specific graph mechanism.

\section{Conclusion}
\label{sec:conclusion}

Long-term memory QA requires deciding what to retrieve next given what has already been found. \methodname{} separates reusable global memory from a question-specific, verifier-gated evidence state and trains a lightweight planner on newly recovered evidence. \methodname{} achieves strong answer accuracy on LoCoMo and LongMemEval-S with both Qwen3-30B and GPT-4o-mini backbones. With Qwen3-30B for evidence processing and answering, its trained 0.6B planner also surpasses a 30B planner without retrieval-grounded training in answer accuracy, while later retrieval rounds recover evidence missed initially. These results support evidence-guided retrieval for long-term memory QA and suggest that, under the evaluated settings, training a lightweight planner can be more effective than increasing planner size without retrieval-grounded training.

\subsection*{AI Use Statement}

We used generative AI tools to polish drafts written by the authors, including improving grammar, phrasing, clarity, and the flow of the text, and checking consistency across sections. We also used AI assistance to refine the visual presentation of figures and to adjust manuscript formatting and layout. For literature retrieval and discovery, we used AI tools to help find additional studies addressing similar problems or using related methods; these works are discussed in the Related Work section. A code agent assisted with writing and debugging experiment code.

The research ideas, method design, and experimental design are entirely our own; no generative AI tool was used to generate research ideas. We have reviewed all AI-assisted content: the manuscript text was checked paragraph by paragraph by the authors for accuracy and clarity, and AI-assisted code was verified by the authors and tested in our actual experiments. We take full responsibility for the final content of this work, including the text, experimental claims, and any artifacts produced with the aid of generative AI tools.

\subsection*{Ethics Statement}

This work uses LoCoMo and LongMemEval-S, two publicly released long-term conversational question-answering benchmarks whose dialogues are synthetic or de-identified and do not contain identifiable private information about real users. This study does not involve new human-subject experiments and therefore did not require institutional review board approval. We note that a retrieval-augmented long-term memory system deployed in a real product would need to carefully address user conversation privacy and controllable management or deletion of stored memories; this paper, however, focuses on the retrieval and planning algorithm itself and does not address a specific product deployment. We are not aware of any funding conflicts, discriminatory bias, or other issues in this work that would raise concerns under the Code of Ethics.

\subsection*{Reproducibility Statement}

We will publicly release the complete code repository and training data once this paper has passed peer review, so that our experimental results can be independently reproduced. The complete implementation details, hyperparameter configurations, and the retrieval-verification procedure are given in Section~\ref{sec:method} and the appendix; complete dataset descriptions, metric accounting, and experimental settings are given in Appendix~\ref{app:experimental-setup}; and the construction of retrieval-grounded policy-training data is detailed in Appendix~\ref{app:rl-data-construction}. All backbone models used in this work are publicly available: Qwen3-30B and Qwen3-0.6B are open-source models, GPT-4o-mini is accessed through a public API, and the embedding model used for retrieval is likewise publicly released, facilitating reproduction of our results under the same conditions.

\bibliography{iclr2027_conference}
\bibliographystyle{iclr2027_conference}

\clearpage
\appendix
\section*{Appendix}

\section{Method Details}
\label{app:method-details}

\subsection{Atomic-Fact Memory and Parent-Turn Fact Aggregation}
\label{app:atomic-memory}

For each conversation turn, an extraction model produces self-contained atomic facts together with keywords and topic tags. Speaker identity, timestamp, and turn identifier are retained as metadata. Pronouns are replaced with explicit entities whenever possible, and subject--action--object directionality is preserved to reduce ambiguity outside the original context. Each fact is embedded together with its keywords and topics, while its parent identifier and the full set of facts extracted from that turn are stored alongside the vector. This memory index is constructed offline and reused across questions associated with the same conversation history.

Matching individual facts provides a focused retrieval signal, while related facts extracted from the same turn can supply additional information for verification. \methodname{} therefore uses fine-to-coarse parent--child retrieval: atomic facts are the matching units, while parent-level fact records are the verification and reasoning units. Each record contains all extracted facts from one turn together with speaker and timestamp metadata. The source-turn identifier preserves provenance; the text presented to the verifier is the extracted fact set, not the verbatim conversation turn.

For each retrieval intent, \methodname{} retrieves the top-$k$ atomic candidates, merges results across intents, and deduplicates them by parent turn. When several facts from the same turn are retrieved, their parent record appears only once and is ranked by its highest atomic matching score. The verifier receives the aggregated fact record with speaker and temporal metadata. Viewed fact and parent-turn identifiers are added to an exclusion set so that the same evidence does not repeatedly enter later verification rounds.

\subsection{Evidence-Graph Construction and Normalization}
\label{app:evidence-graph}

The graph implements the structured record within the question-specific evidence state. In $G_t=(V_t,L_t,X_t,D_t,\Psi_t)$, $V_t$ contains typed nodes for people, organizations, locations, events, dates, quantities, and states. $L_t$ is a set of directed labeled relations; $X_t$ records active entities that can anchor further retrieval; $D_t$ stores search directions invalidated or exhausted by the current evidence; and $\Psi_t$ maps graph elements to their supporting parent turns. The complete evidence state $s_t=(q,G_t,Q_t,S_t,\mathcal{E}_t)$ also contains previous queries $Q_t$, viewed candidates $S_t$, and accumulated accepted evidence $\mathcal{E}_t$. These records guide the next query; they do not define the contents or connectivity of the searchable memory.

Before retrieval, the evidence model extracts a small set of salient people, events, locations, or concepts from the question as seed nodes. If structured parsing fails, \methodname{} uses the complete question as a fallback node. In later rounds, the verifier converts only directly useful candidates into entities, attributes, and labeled relations, retaining provenance for every new node and edge. Graph merging reuses matching entities, removes duplicate relations, and assigns versioned identifiers to conflicting attributes so that states from different times can coexist. A direction is marked invalid only when new evidence explicitly contradicts it or establishes that it no longer holds.

A conversation timestamp indicates when a dialogue turn occurred, not necessarily when the described event occurred. The verifier therefore distinguishes turn time from event time: exact relative expressions are resolved against the turn timestamp, vague expressions retain anchored intervals, and the turn timestamp is not used as event time when no temporal expression is present~\cite{TimeML,HeidelTime}. The same explicit calculation process handles ages, amounts, counts, and durations. Normalized values are stored with their original statements and provenance.

\subsection{Retrieval Execution, Verification, and Termination}
\label{app:active-retrieval}

Each structured intent $I_t^\ell=(d_t^\ell,k_t^\ell,c_t^\ell)$ specifies the desired fact, key entities or concepts, and topical labels. The planner compares the requirements of the question with the evidence state to identify missing entities, relations, dates, or quantities. Previous queries are explicitly included in the state, encouraging alternative expressions and search directions when earlier actions have not resolved a gap. The retriever concatenates the intent fields into query text and searches the global atomic-fact index, excluding previously viewed evidence. It does not traverse the evidence graph to enumerate candidates, so retrieved facts can introduce entities not yet recorded in that graph.

After the retriever returns parent-level fact records, the evidence model verifies each candidate using the question, current graph, and previously accepted evidence. Records judged useful for resolving the question enter the accepted set. Subsequent planning therefore depends on verified evidence rather than unfiltered retrieval text.

The loop terminates when the planner or verifier judges the evidence sufficient after a minimum exploration requirement, when several consecutive rounds yield no valid intent, candidate, or graph update, or when the process reaches $T_{\max}$. The answer model then generates an answer from the question, final graph, and accumulated accepted fact records. If retrieval stops because of exhaustion or the round limit, the model answers only from the currently verified evidence.

\subsection{Planner Training Details}
\label{app:policy-optimization}

\subsubsection{Training Data Construction}
\label{app:rl-data-construction}

For LoCoMo, we use the data-generation utilities released in the benchmark repository. The generator samples detailed personas and dated personal events, organizes those events into multi-session schedules, and produces long conversations conditioned on the resulting profiles and timelines. A question generator then reads each complete conversation and creates questions together with answers and supporting dialogue turns. This procedure provides many distinct conversational histories while preserving the persona, event, and temporal structure needed for long-term memory retrieval.

For LongMemEval, we use conversations from our training partition, defined in Appendix~\ref{app:experimental-setup}, rather than synthesizing new histories. We generate additional questions from each training conversation, require every answer to be grounded in the conversation, and align the supporting evidence with retrievable parent turns. The two procedures therefore differ only in how the underlying questions and conversations are obtained: LoCoMo supplies a repository-native pipeline for generating both, whereas LongMemEval supplies the training conversations and \methodname{} augments them with additional questions.

Table~\ref{tab:rl-data-construction} summarizes the resulting data. The LoCoMo pipeline generates both the conversation histories and their questions. For LongMemEval, the question count combines the two augmentation rounds used by the final training set. Its collected-state count includes all trajectory states from these rounds; evidence validation and turn alignment are applied before rollout filtering, which reduces the number of states entering that stage.

\begin{table}[t]
\caption{Construction and filtering of planner-training data. A state is the input saved before one planner call in a retrieval trajectory. Retention is measured against the states entering rollout filtering.}
\label{tab:rl-data-construction}
\centering
\small
\setlength{\tabcolsep}{3.5pt}
\begin{tabular}{lrrrrr}
\toprule
Dataset & Histories & Generated QAs & Collected states & Rollout input & Retained \\
\midrule
LoCoMo & 170 & 24,787 & 74,742 & 74,742 & 2,996 (4.0\%) \\
LongMemEval & 277 & 11,017 & 17,752 & 14,475 & 2,312 (16.0\%) \\
\bottomrule
\end{tabular}
\end{table}

\subsubsection{Multi-Round State Collection and Filtering}

After question construction, both datasets use the same trajectory collection and filtering procedure. Training states are collected from complete \methodname{} trajectories. Before every planner call, the system stores the question, evidence graph, query history, complete gold evidence set $\mathcal{E}^*$, gold evidence already found $F_t$, and the corresponding memory. The data therefore cover initial retrieval, complementary retrieval from partial evidence, and late-stage evidence completion.

For each state, the base planner generates multiple rollouts whose queries are executed against the actual memory. A state is too difficult for the current policy when every rollout retrieves no new evidence, and too easy when every rollout retrieves all remaining evidence. \methodname{} removes both cases and retains states with mixed or partial outcomes, concentrating training on decisions for which policy choice can change the retrieval result.

\subsubsection{Incremental Turn-Level Retrieval Reward}

The retrieval reward is defined in Section~\ref{sec:policy-optimization}. It evaluates whether a query finds evidence still missing from the current state. Evidence already found receives no additional reward. At the beginning of retrieval, $R_{\mathrm{ret}}$ reduces to recall over the remaining evidence. As search progresses, the progress term increases the reward for partial recovery of the remaining evidence. Reward is computed at the parent-turn level, matching the evidence unit used by verification and inference.

The reward has five useful properties. First, it provides action-level feedback immediately after query execution, shortening the credit-assignment path compared with final-answer reward. Answer correctness is jointly affected by evidence verification, temporal calculation, and answer generation, whereas newly recovered evidence more directly measures the quality of the current query. Second, the reward evaluates retrieval outcomes rather than similarity to a prescribed query string. Multiple query formulations receive equivalent feedback when they recover the same target evidence, which is important because effective queries are not unique.

Third, only uncovered evidence contributes to reward. Repeatedly retrieving an existing turn receives no additional credit, shifting the objective from maximizing raw hits to reducing the current evidence gap and encouraging complementary actions across rounds. Fourth, for a fixed partial-recovery fraction $0<x_t<1$, the progress term $p_t(1-x_t)$ gives additional weight to actions taken after more gold evidence has already been found. This term encourages continued evidence acquisition from incomplete states later in a trajectory. Fifth, reward is computed at the parent-turn level, the same unit used for gold annotations, evidence verification, and answer generation. This alignment reduces a granularity mismatch between training and inference.

The reward complements informative-state filtering. Filtering first removes states in which policy choice cannot change the outcome, and incremental reward then distinguishes how much the remaining actions reduce the evidence gap. Training computation is therefore concentrated on decision points with meaningful policy variation rather than diluted by uniformly successful or uniformly unsuccessful samples.

We additionally regularize planner outputs with a keyword-count penalty that discourages overly broad retrieval intents and a response-length penalty that discourages excessively short or verbose actions. These auxiliary terms shape the form of the retrieval action, while the executed retrieval outcome remains the primary learning signal.

\subsubsection{Policy Optimization}

We train the planner with Group Relative Policy Optimization (GRPO). Unlike actor--critic methods, GRPO does not fit a separate value model. For a planner state $s$, the old policy $\pi_{\mathrm{old}}$ samples a group of $G$ responses $o_1,\ldots,o_G$. \methodname{} parses and executes every response against the same memory and assigns it a scalar reward $r_i$. The group mean and standard deviation define the relative advantage

\begin{equation}
\bar r=\frac{1}{G}\sum_{i=1}^{G}r_i,\qquad
\sigma_r=\sqrt{\frac{1}{G-1}\sum_{i=1}^{G}(r_i-\bar r)^2},\qquad
A_i=\frac{r_i-\bar r}{\sigma_r+\epsilon}.
\end{equation}

Thus, an output receives a positive advantage only when it performs better than other outputs generated for the same evidence state. This within-group comparison controls for variation in question and state difficulty without requiring a learned critic. The same sequence-level advantage $A_i$ is applied to every generated token in $o_i$. For token $o_{i,j}$, let

\begin{equation}
\rho_{i,j}(\theta)=
\frac{\pi_\theta(o_{i,j}\mid s,o_{i,<j})}
{\pi_{\mathrm{old}}(o_{i,j}\mid s,o_{i,<j})}.
\end{equation}

GRPO maximizes the token-aggregated clipped surrogate

\begin{equation}
J_{\mathrm{clip}}(\theta)=
\frac{1}{\sum_i |o_i|}
\sum_{i=1}^{G}\sum_{j=1}^{|o_i|}
\min\!\left(
\rho_{i,j}(\theta)A_i,
\operatorname{clip}\!\left(\rho_{i,j}(\theta),1-\varepsilon,1+\varepsilon\right)A_i
\right).
\end{equation}

Clipping limits destructive policy updates when the new policy moves too far from the rollout policy. We additionally regularize the planner toward a fixed reference policy and optionally encourage exploration through token entropy:

\begin{equation}
J(\theta)=J_{\mathrm{clip}}(\theta)
-\beta D_{\mathrm{KL}}\!\left(\pi_\theta\,\|\,\pi_{\mathrm{ref}}\right)
+\eta\,\mathcal{H}(\pi_\theta),
\end{equation}

where $\beta$ and $\eta$ control the KL and entropy terms, respectively, and $\eta$ may be zero. Because all members of a group share the question, graph, query history, and memory, reward differences primarily measure the quality of their query actions. If every response in a group obtains the same reward, its normalized advantages vanish and the state contributes no comparative learning signal. The informative-state filtering described above removes states that the base planner solves uniformly or fails uniformly, increasing the frequency of groups with meaningful reward variation. GRPO then shifts probability mass toward executable query actions that recover more of the missing evidence while constraining policy drift.

\subsubsection{Scope and Future Directions of RL Training}

Our RL design is deliberately simple: a lightweight planner is trained with GRPO using incremental evidence recovery as the main reward, while the memory index, evidence model, and answer model remain fixed. The trained-versus-untrained comparison supports the effectiveness of this training setup as a whole.

Future work could compare reward variants under matched training data and compute budgets, including evidence recall alone, incremental recall without progress weighting, and combinations of retrieval feedback and final-answer correctness. Ablating the auxiliary penalties and state filtering separately would help identify which components contribute to the observed gains. Beyond these comparisons, cost-aware rewards could balance evidence acquisition against retrieval rounds and token use, while trajectory-level optimization could jointly train query selection and stopping decisions. These extensions would examine whether improvements in immediate evidence recovery also translate into more efficient and accurate multi-round reasoning.

\section{Expanded Experimental Setup}
\label{app:experimental-setup}

\subsection{Datasets}

\paragraph{LoCoMo.}
LoCoMo contains 10 long conversations and 1,986 questions. Each conversation records two speakers interacting across multiple time points and includes personal events, relations, preferences, and evolving states. The five question categories are multi-hop, temporal, open-domain, single-hop, and adversarial. LoCoMo therefore evaluates factual retention, cross-session evidence composition, temporal reasoning, and responses to questions whose answers are unsupported by the conversation.

\paragraph{LongMemEval-S.}
LongMemEval-S contains 500 questions for evaluating whether conversational assistants can retain, retrieve, and use information from long interaction histories. Each question is associated with multiple historical sessions, of which only a small subset supports the answer. Its categories include single-session user and assistant information, user preferences, multi-session integration, temporal reasoning, and knowledge updates. The benchmark emphasizes sparse evidence localization under substantial historical distraction.

The datasets are complementary: LoCoMo emphasizes information evolution in continuous conversations, whereas LongMemEval-S emphasizes sparse evidence in high-noise histories. We evaluate answer accuracy on all 1,986 LoCoMo questions. All LongMemEval-S test evaluations reported in this paper use the same held-out test split of 101 questions, including the main comparisons, ablations, and supplementary analyses. According to the data split, instances outside the test set are used for data generation, as described in Appendix~\ref{app:rl-data-construction}.

\subsection{Compared Methods}

The main comparison includes FullText, NaiveRAG, IRCoT, GraphReader, LangMem, A-MEM, MemoryOS, Mem0, and LightMem. FullText directly supplies the complete history to the answer model, while NaiveRAG performs one-shot similarity retrieval. IRCoT and GraphReader provide iterative retrieval comparisons. The remaining systems represent structured, dynamically updated, or compressed long-term memory. These comparisons examine both persistent memory design and how retrieval proceeds while answering a question. We report results with Qwen3-30B-A3B-Instruct-2507 and GPT-4o-mini as the backbone.

Unless stated otherwise, we use Qwen3-30B-A3B-Instruct-2507 for cache construction, seed-graph generation, evidence verification and graph update, and final answer generation. In the GPT-4o-mini experiments, we rebuild the cache with GPT-4o-mini and use it for the same inference stages while holding the planner configuration fixed.

\subsection{Metrics}

\paragraph{Answer accuracy.}
We use LLM-as-a-Judge Accuracy as the primary task metric. The judge reads the question, reference answer, and model answer and determines whether the response is correct. Accuracy is the proportion of correctly answered questions in the evaluation set.

\paragraph{Generative-model cost.}
We report generative-model token usage separately for offline memory construction and online question answering. Summary In and Summary Out measure the input and output tokens used to transform raw conversations into retrievable memories. Update In and Update Out measure persistent memory-update tokens. Their sum is reported as Total. \methodname{} performs atomic-fact extraction but no separate persistent update, so its Update entries are denoted by ``--'' rather than zero.

LongMemEval-S provides a separate history for each question, so offline costs are averaged per question (k/Q). In LoCoMo, one memory is shared by all questions associated with a conversation, so offline costs are averaged over the 10 conversations (k/Conv.). Online In, Out, and Total are averaged per question for both datasets. Embedding and vector retrieval do not invoke a generative model and are excluded from token counts.

\subsection{Implementation Details}

\methodname{} uses a query planner to generate retrieval intents from the current evidence state and BAAI/bge-large-en-v1.5 to embed atomic facts and retrieval intents. The main results on both datasets use \methodname{}-RL, with the cache-construction and inference backbone set as described above.

We adapt IRCoT and GraphReader to the same conversational question sets using Qwen3-30B-A3B-Instruct-2507 for generation and judging. IRCoT retrieves six turns per BM25 query, accumulates at most 15 passages of up to 350 words each, and generates at most ten reasoning sentences at temperature zero. GraphReader uses session-dated histories, sentence-aligned chunks of 1,000 characters for LoCoMo and 4,000 for LongMemEval-S, and local BGE-large-en-v1.5 embeddings. Its fact-exploration and chunk-reading agents allow at most 12 and 16 calls, respectively, at temperature 0.4. The larger LongMemEval-S chunk size is a resource adjustment in this adaptation.

For these two baselines, costs use the input and output usage returned by each generative API call, including generated text discarded by subsequent processing and logical prompt tokens served from a prefix cache. Judge calls are excluded. Exact replay of saved responses during recovery is counted once; interrupted requests that returned no usage are not measurable and are excluded. Neither baseline performs a separate persistent memory update.

\section{Online Cost, Runtime, and Termination Diagnostics}
\label{app:online-diagnostics}

\subsection{Online Inference Cost}
\label{app:online-cost}

Tables~\ref{tab:online-cost-locomo} and~\ref{tab:online-cost-lme} report the input and output token costs underlying Section~\ref{sec:online-cost}, using Qwen3-30B throughout. Both datasets use thousands of tokens per question (k/Q), excluding offline construction, embeddings, and judging. \methodname{}'s online cost includes seed-graph construction, query planning, evidence verification and graph update, and final answer generation.

\begin{table}[t]
\caption{Online inference cost on LoCoMo, in thousands of tokens per question (k/Q).}
\label{tab:online-cost-locomo}
\centering
\begin{tabular}{lrrr}
\hline
\textbf{Method} & \textbf{In (k)} & \textbf{Out (k)} & \textbf{Total (k)} \\
\hline
IRCoT & 4.29 & 0.13 & 4.43 \\
GraphReader & 24.50 & 0.54 & 25.04 \\
\methodname{} & 13.47 & 1.80 & 15.27 \\
\hline
\end{tabular}
\end{table}

\begin{table}[t]
\caption{Online inference cost on LongMemEval-S, in thousands of tokens per question (k/Q).}
\label{tab:online-cost-lme}
\centering
\begin{tabular}{lrrr}
\hline
\textbf{Method} & \textbf{In (k)} & \textbf{Out (k)} & \textbf{Total (k)} \\
\hline
IRCoT & 19.02 & 0.24 & 19.26 \\
GraphReader & 32.82 & 0.52 & 33.34 \\
\methodname{} & 14.97 & 2.43 & 17.40 \\
\hline
\end{tabular}
\end{table}

\subsection{Runtime Statistics}
\label{app:runtime-statistics}

We measure online runtime on all 101 LongMemEval-S test questions using a shared prebuilt memory cache on an NVIDIA GeForce RTX 4090 server. Qwen3-30B-A3B-Instruct-2507 is served by SGLang with tensor parallelism across four GPUs, while Qwen3-0.6B is served by a separate SGLang instance on one GPU. The evaluation uses sample-level parallelism of 32. Runtime per question includes query planning, retrieval, evidence verification and graph update, and final answer generation.

\begin{table}[t]
\caption{Online runtime on LongMemEval-S. Calls and wall-clock time are averaged per question and then across three \methodname{}-Small runs, two \methodname{}-RL runs, and three \methodname{}-Large runs. Relative runtime is normalized by \methodname{}-Small.}
\label{tab:runtime-statistics}
\centering
\begin{tabular*}{\textwidth}{@{\extracolsep{\fill}}lccc@{}}
\hline
\textbf{Configuration} & \textbf{LLM calls/Q} & \textbf{Runtime/Q (s)} & \textbf{Relative runtime} \\
\hline
\methodname{}-Small & 6.8 & 43.2 & 1.00$\times$ \\
\methodname{}-RL & 6.6 & 48.3 & 1.12$\times$ \\
\methodname{}-Large & 10.9 & 155.5 & 3.60$\times$ \\
\hline
\end{tabular*}
\end{table}

Policy training increases runtime by only 5.1 seconds per question because \methodname{}-RL retains the same 0.6B planner architecture and serving setup as \methodname{}-Small. \methodname{}-Large is substantially slower because every planning step invokes the 30B service and its measured trajectories contain more LLM calls. These results show that retrieval-grounded training improves the behavior of the lightweight planner without incurring the inference cost of replacing it with the large backbone.

\subsection{Termination Behavior}
\label{app:termination-behavior}

\begin{figure}[t]
\centering
\includegraphics[width=\linewidth]{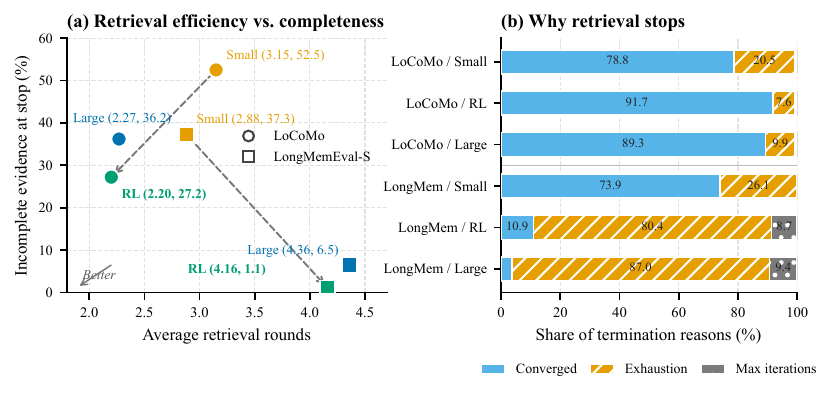}
\caption{Termination efficiency and stopping mechanisms. Left: average retrieval rounds versus the fraction of questions terminating before all annotated gold turns are recovered; lower-left is better. Within each dataset, the dashed arrow from \methodname{}-Small to \methodname{}-RL shows the change in overall system behavior after reinforcement learning. Right: the complete distribution of convergence, retrieval exhaustion, and maximum-iteration stops. Recall below one is an upper-bound diagnostic because partial gold evidence may already suffice for some answers.}
\label{fig:termination-behavior}
\end{figure}

Figure~\ref{fig:termination-behavior} shows that most questions terminate well before the maximum retrieval budget. On LoCoMo, \methodname{}-Small averages 3.15 rounds yet terminates without complete gold evidence on 52.5\% of questions. \methodname{}-RL occupies a substantially better operating point, with 2.20 rounds and 27.2\% incomplete recall, while \methodname{}-Large reaches 2.27 rounds and 36.2\%. For \methodname{}-Small, verifier-declared convergence accounts for 78.8\% of stops, while retrieval exhaustion accounts for 20.5\%. These stopping frequencies accompany a final evidence recall of 0.514. They motivate examining whether continued retrieval could recover useful evidence, but neither incomplete gold recall nor the aggregate stopping frequencies establish that a particular sufficiency decision was incorrect: some questions can be answered from a subset of the annotated turns.

On LongMemEval-S, the \methodname{}-RL run analyzed here reduces the incomplete-recall rate from 37.3\% for \methodname{}-Small to 1.1\%, while increasing average rounds from 2.88 to 4.16. The shift is downward but to the right: the stronger result combines effective early retrieval (a mean first-hit round of 1.13) with continued search for missing evidence, rather than obtaining completeness by stopping earlier. \methodname{}-RL terminates through retrieval exhaustion in 80.4\% of cases, reaches the maximum iteration limit in 8.7\%, and stops through verifier-declared convergence in 10.9\%. Compared with \methodname{}-Large, it achieves lower incomplete recall (1.1\% versus 6.5\%) with slightly fewer rounds (4.16 versus 4.36), and therefore dominates \methodname{}-Large on these two diagnostic axes. It nevertheless uses more retrieval rounds than \methodname{}-Small. These results diagnose efficiency and stopping mechanisms rather than constitute a controlled stopping-policy ablation.

\section{Planner Behavior and Evidence Packaging}
\label{app:planner-diagnostics}

\subsection{Retrieval Depth and Retriever Compatibility}
\label{app:retrieval-depth-interface}

\paragraph{Evidence gain diminishes with retrieval depth.}
At least 92\% of each LoCoMo curve's post-first-round recall improvement is realized by round 5, after which the curves change by at most 1.9 points. High-similarity evidence is recovered first, leaving progressively weaker matches and more indirect relations. This diminishing evidence gain motivates adaptive termination rather than mechanically spending the maximum retrieval budget.

\paragraph{Planner outputs must match the retriever.}
Under dense retrieval on LoCoMo, \methodname{}-RL exceeds \methodname{}-Small by 18.2 points in the first round and by 29.1 points at termination. \methodname{}-Small gains only 2.3 points after round 3, whereas \methodname{}-RL gains a further 6.1 points, indicating that policy training improves both initial query quality and the ability to use accumulated evidence for follow-up retrieval. The earlier keyword diagnostic in Table~\ref{tab:query-behavior} shows a much smaller RL--Small gap, indicating that planner gains depend on how well the learned query policy matches the retriever.

\subsection{Graph-Structured Versus Flat Evidence}
\label{app:graph-ablation}

\paragraph{Configuration.}
The no-graph ablation uses dense atomic-fact retrieval with MMR ($\lambda=0.95$), $k=8$, and a maximum of 10 retrieval rounds. Parent-turn fact aggregation remains enabled. Qwen3-30B-A3B-Instruct-2507 performs evidence verification and final answer generation. The planners are Qwen3-0.6B without retrieval-grounded training, retrieval-trained Qwen3-0.6B, and Qwen3-30B without retrieval-grounded training. Figure~\ref{fig:graph-ablation} reports all six no-graph results alongside their graph references on the same question sets used elsewhere in the paper.


\paragraph{Intervention.}
The flat-evidence variant removes the evidence graph and its associated construction, update, and state-maintenance mechanisms. Accepted evidence accumulates across rounds and remains available to the planner and final answer model as a flat list.

\paragraph{Results.}
All six graph--flat gaps are positive, with graph accuracies taken from the corresponding configurations in Table~\ref{tab:scale-rl-ablation}. For \methodname{}-Small, \methodname{}-RL, and \methodname{}-Large, the gaps are 14.02, 14.41, and 9.65 points on LoCoMo and 5.34, 16.17, and 7.26 points on LongMemEval-S, respectively.


\subsection{Changes in Query Behavior}
\label{app:query-behavior}

\begin{table}[t]
\caption{LoCoMo query behavior and retrieval outcomes before and after retrieval-grounded training. The comparison between \methodname{}-Small and \methodname{}-RL examines how policy training changes the use of the retrieval interface: query word count and intents per call characterize action construction; repetition and parsing failures characterize action quality; and evidence recall and premature stopping measure downstream retrieval effectiveness. Change is computed as \methodname{}-RL minus \methodname{}-Small.}
\label{tab:query-behavior}
\begin{center}
\begin{tabular}{lrrr}
\hline
Metric & \methodname{}-Small & \methodname{}-RL & Change \\
\hline
Average query word count & 3.30 & 3.50 & +0.20 \\
Intents per planner call & 2.75 & 2.98 & +0.23 \\
Normalized query repetition & 17.1\% & 16.0\% & $-1.1$ pp \\
JSON parsing failures & 0.0\% & 0.0\% & 0.0 pp \\
First-round evidence recall & 0.623 & 0.631 & +0.8 pp \\
Final evidence recall & 0.761 & 0.776 & +1.5 pp \\
Premature stopping & 29.0\% & 27.2\% & $-1.8$ pp \\
\hline
\end{tabular}
\end{center}
\end{table}

Both planners produce parseable output, so training does not merely repair a formatting failure. Query length increases by only 0.20 words, while the number of intents rises modestly and normalized repetition falls by 1.1 points. Final evidence recall and premature stopping improve in the same direction. These changes are consistent with the incremental reward: repeated evidence receives no additional credit, whereas complementary retrieval actions receive immediate feedback. The present statistics do not isolate the independent contributions of intent count, keyword choice, and stopping behavior.

\subsection{A Trajectory from the RL-Trained Planner}
\label{app:rl-trajectory}

We illustrate how accumulated evidence informs what to retrieve next using a recorded LongMemEval-S test trajectory (question ID: \texttt{gpt4\_5501fe77}). The planner is a GRPO-trained Qwen3-0.6B model. The question is: \emph{``Which social media platform did I gain the most followers on over the past month?''} Neither Twitter, Facebook, nor TikTok is named in the question. We show consecutive rounds 2--4 below, with their preceding evidence state and the eventual answer. Quoted queries and fact excerpts retain the wording in the log; graph descriptions summarize the relevant nodes and relations.

\paragraph{Evidence available after round 1.}
The initial graph contains only the generic anchors \emph{social media platform}, \emph{followers growth}, and \emph{past month}. The first retrieval batch supplies two useful comparisons. \texttt{Evidence\_003}, dated May 29, 2023, states: ``The user observed that their Twitter follower count increased from 420 to 540 over the past month.'' \texttt{Evidence\_004}, dated May 30, states: ``The user has observed that their Facebook follower count has remained stable at approximately 800.'' The verifier adds Twitter and Facebook as active entities, with relations to attributes recording 120 followers gained and a stable follower count, respectively. TikTok is mentioned incidentally in some retrieved records, but has no node or accepted growth evidence in the graph at this stage.

\paragraph{Round 2: querying verified anchors.}
The planner now issues ``What is the number of followers on Twitter over the past month?'' and ``What is the number of followers on Facebook over the past month?'' It also retains a broader comparison query: ``Which social media platform had the highest follower increase during the past month?'' The named queries use entities established in the preceding graph. This round adds no new platform entity, and TikTok remains absent from the graph.

\paragraph{Round 3: retrieving beyond the current graph.}
The next batch contains ``How many followers did Twitter gain in the past month?'', ``How many followers did Facebook gain in the past month?'', and ``Which social media platform had the highest follower increase during the past month?'' Retrieval returns \texttt{Evidence\_025}, dated May 29: ``The user gained approximately 200 followers on TikTok over the past three weeks.'' It also returns \texttt{Evidence\_028}, dated May 30: ``The user has observed a rapid increase in their TikTok follower count.'' The verifier accepts both records, creates an active TikTok entity, and links it through the \texttt{followers\_growth\_in\_period} relation to the newly supported growth attributes. The quantitative evidence is accessible even though TikTok is not yet a node in the graph supplied to this round's planner. The log records the merged retrieval batch, so we do not attribute either record to an individual intent.

\paragraph{Round 4: using the newly admitted entity.}
The next planner call includes ``What is the number of followers on TikTok over the past month?'' alongside corresponding queries for Twitter and Facebook. TikTok has thus moved from a retrieved fact into the verified graph and then into a subsequent retrieval intent. Round 4 adds no new accepted evidence IDs; the verifier revisits the existing growth records. This step illustrates how the graph conditions the next action, without implying that every such action yields additional evidence.

\paragraph{Answer and interpretation.}
The full run ends after six rounds with the logged termination reason \texttt{exhaustion}, and returns ``TikTok'', matching the benchmark reference. This selected trajectory makes the mechanism concrete: verified graph entities provide query anchors, global retrieval admits relevant evidence about an entity outside the current graph, and graph updates make that entity available to subsequent planning. It illustrates the behavior of an RL-trained planner; it does not by itself isolate the contribution of RL or establish an efficiency advantage.

\subsection{Parent-Turn Fact Aggregation}
\label{app:parent-turn-ablation}
\label{sec:parent-turn-ablation}

\methodname{} performs matching over atomic facts and groups each match with all other facts extracted from the same parent turn before verification. This design aims to combine fine-grained matching with additional facts associated with the retrieved turn. We ablate parent-turn fact aggregation on all 1,986 LoCoMo questions: each matched atomic fact is returned directly as a separate candidate. Both variants retain speaker and timestamp metadata. Within each planner row, the memory cache, code version, retrieval parameters, answer model, and judge are held fixed. The analysis therefore concerns the paired change within each row rather than absolute accuracy differences across planners.

Removing parent-turn fact aggregation changes answer accuracy by $-0.46$ points for \methodname{}-RL, $-2.77$ points for \methodname{}-Small, and $-4.66$ points for \methodname{}-Large. All three paired comparisons favor aggregation, with larger drops for the two planners without retrieval-grounded training. These results are consistent with sibling facts providing useful information beyond the matched fact. The comparison does not test access to the original wording of the turn, since both variants present extracted facts.

One plausible explanation for \methodname{}-RL's smaller drop is that retrieval-grounded training encourages queries whose matched facts more directly address the question, reducing reliance on additional facts from the same parent turn. The reward favors newly recovered evidence, and Table~\ref{tab:query-behavior} reports slightly more intents, less repetition, and higher final evidence recall after training. These observations are compatible with more targeted retrieval, but do not directly measure the sufficiency of individual matched facts or establish why \methodname{}-RL is less sensitive to aggregation.

The ablation also changes how evidence is packaged. Without aggregation, the verifier receives shorter fact records, while sibling facts from one source turn can occupy separate candidate slots despite a fixed candidate budget. The result therefore evaluates the aggregation interface as a whole; it does not isolate the contributions of additional fact content, input length, or candidate competition.

\section{Question Types and Failure Attribution}
\label{app:error-analysis}
\label{sec:category-analysis}

\subsection{Accuracy by Question Type}

\begin{figure}[t]
\centering
\includegraphics[width=0.49\linewidth]{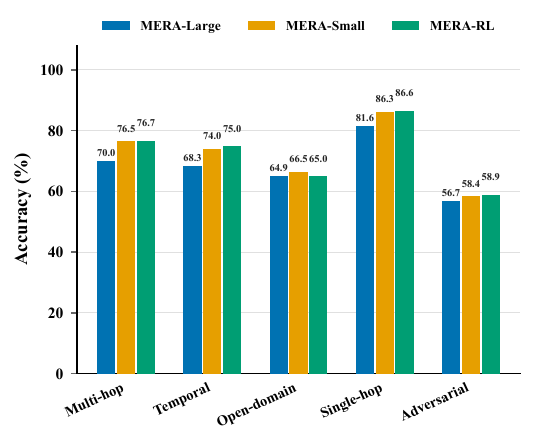}\hfill
\includegraphics[width=0.49\linewidth]{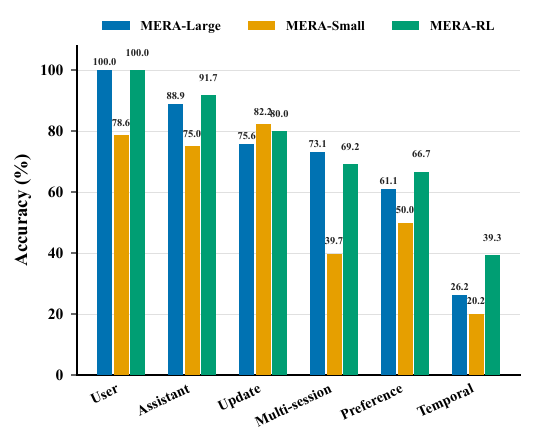}
\caption{Accuracy by question category on (left) LoCoMo and (right) LongMemEval-S. Values above bars are accuracies (\%).}
\label{fig:category-accuracy}
\end{figure}

LoCoMo shows a consistent category ordering across planners: single-hop accuracy is highest, followed by multi-hop, temporal, open-domain, and adversarial accuracy. The high single-hop scores are consistent with questions that can be resolved from a localized fact. Adversarial questions test whether a model avoids unsupported answers, so their lower accuracy should be interpreted as difficulty handling absent information rather than updating stored knowledge. Open-domain questions can require knowledge beyond the conversation, making their scores less directly attributable to conversational evidence retrieval alone. Differences between \methodname{}-Small and \methodname{}-RL remain within 1.5 points for every category, consistent with the small overall improvement in the matched keyword setting.

The largest descriptive difference between \methodname{}-Small and \methodname{}-RL on LongMemEval-S occurs for multi-session questions (29.5 points), followed by single-session user information (21.4 points) and temporal questions (19.1 points). Multi-session questions naturally expose decomposable information needs, allowing evidence from one session to anchor retrieval in another. Temporal accuracy remains only 39.3\%, however, showing that making temporal evidence visible does not guarantee correct normalization, ordering, or arithmetic.

\subsection{Failure Attribution}
\label{app:failure-attribution}

\begin{figure}[t]
\centering
\includegraphics[width=\linewidth]{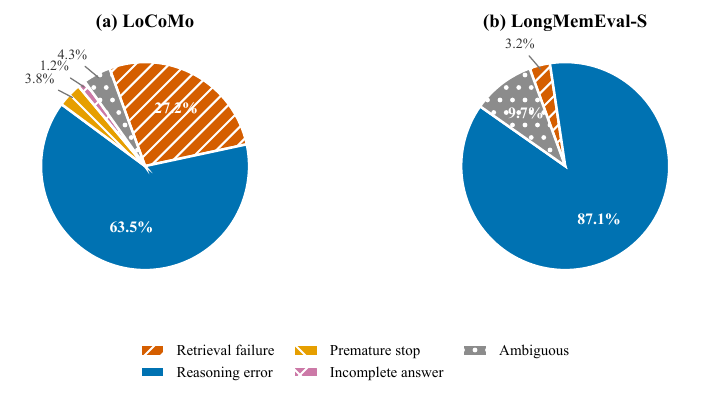}
\caption{Attribution of incorrect \methodname{}-RL answers on LoCoMo and LongMemEval-S. ``Reasoning error'' denotes an incorrect final answer despite recovered evidence; it localizes the failed stage but is not itself a fine-grained root cause. Percentages are diagnostic labels over errors rather than manually adjudicated absolute rates.}
\label{fig:failure-attribution}
\end{figure}

Long-term memory QA can be approximated as

\begin{equation}
P(\text{correct})\approx
P(\text{evidence sufficient})
P(\text{correct}\mid\text{evidence sufficient}).
\end{equation}

The attribution analysis covers 655 incorrect LoCoMo answers and 31 incorrect LongMemEval-S answers. Reasoning errors are the largest category on both datasets, accounting for 63.5\% of LoCoMo failures and 87.1\% of LongMemEval-S failures. This category provides a coarse stage-level diagnosis: it shows that the system found supporting evidence but still produced an incorrect answer. Identifying the root cause requires a second-level breakdown by question and error type.

The LongMemEval-S breakdown exposes three distinct causes. First, temporal-reasoning questions contribute 40--60\% of reasoning errors across configurations; in the \methodname{}-RL run analyzed here, they account for 14 of 27 cases (51.9\%). Inspection attributes many of these failures to incorrect relative-date calculations, so evidence availability alone does not solve temporal normalization and arithmetic. Second, single-session-user questions contribute no reasoning errors: once the relevant simple fact is retrieved, the answer model responds correctly. Third, multi-session questions contribute a nearly unchanged 7--9 reasoning errors across planner configurations, indicating that cross-session evidence integration remains difficult even as retrieval improves. Reasoning error is therefore an intermediate diagnosis, while temporal calculation and cross-session integration are the more specific downstream root causes.

The balance of retrieval and reasoning errors differs sharply across datasets. Retrieval failure accounts for 27.2\% of LoCoMo errors but only 3.2\% on LongMemEval-S, consistent with the lower final evidence recall on LoCoMo in Figure~\ref{fig:evidence-recall}. A LoCoMo question has approximately three to four gold turns on average, compared with about 1.5 on LongMemEval-S, making complete evidence acquisition more difficult. By contrast, stronger retrieval on LongMemEval-S shifts most remaining errors to downstream reasoning. Premature stopping is uncommon in both runs (3.8\% and 0.0\%), so it is not the primary source of failure in this comparison.

The attribution combines deterministic trajectory checks with local 30B-model judgments. The percentages in Figure~\ref{fig:failure-attribution} should therefore be interpreted as diagnostic evidence about bottlenecks rather than exact causal rates.

\section{Retrieval Hyperparameter Sensitivity}
\label{app:retrieval-sensitivity}

With $k=8$ and $T_{\max}=10$, accuracy on LoCoMo is 75.73\%. Setting $k=4$ or $k=16$ yields 74.97\% and 77.63\%, respectively, while reducing the retrieval budget to five rounds yields 75.88\%. Across these settings, accuracy ranges from 74.97\% to 77.63\%. Increasing retrieval depth to $k=16$ gives the highest observed accuracy, whereas reducing the round budget to five yields accuracy close to the default setting. The latter observation is consistent with the average retrieval length of approximately 2.2 rounds, as most questions terminate before reaching either budget.

\end{document}